\documentclass[letterpaper, 10 pt, journal, twoside]{IEEEtran}  

\usepackage{graphicx}
\usepackage{amsmath}
\usepackage{algorithmic}
\usepackage{algorithm}
\usepackage{amssymb}
\usepackage{censor,xcolor}
\usepackage[table]{xcolor}
\def\censorcolor{gray!50} 
\let\svcensorrule\censorrule \renewcommand\censorrule[1]{ \textcolor{\censorcolor}{\svcensorrule{#1}}}
\usepackage{cite}
\makeatletter
\let\NAT@parse\undefined
\makeatother
\usepackage{hyperref}
\usepackage[switch]{lineno}
\usepackage{soul}
\usepackage{multirow}

\title{
Trajectory and Morphology Planning for Morphing Quadrotors in Ultra-Narrow Passages using Limited Perception
}

\author{Harsh Modi$^{1}$, Xiao Liang$^{2}$ Minghui Zheng$^{1}$
\thanks{\copyright 2026 IEEE. Personal use of this material is permitted.  Permission from IEEE must be obtained for all other uses, in any current or future media, including reprinting/republishing this material for advertising or promotional purposes, creating new collective works, for resale or redistribution to servers or lists, or reuse of any copyrighted component of this work in other works.}
\thanks{This work was partially supported by U.S. National Science Foundation (Grant: No. 2422698). Correspondence to Minghui Zheng.}
\thanks{$^{1}$Harsh Modi {\tt\small harsh.modi@tamu.edu} and Minghui Zheng {\tt\small mhzheng@tamu.edu} are with the Department of Mechanical Engineering, Texas A$\&$M University, College Station, TX 77843 USA}
\thanks{$^{2}$Xiao Liang {\tt\small xliang@tamu.edu} is with the Department of Civil \& Environmental Engineering, Texas A$\&$M University, College Station, TX 77843 USA}}

\begin{document}
\maketitle
\begin{abstract}
This paper presents an online trajectory and morphology planning framework for morphing quadrotors to traverse passages narrower than their unfolded width using only a 2D LiDAR for obstacle sensing. Fixed-frame vehicles cannot pass through such gaps, and classical artificial potential field (APF) based obstacle costs seal them off for planners as well, because the repulsive cost grows sharply inside a tight gap. The core enabler of our framework is a novel cost reduction factor, computed in closed form from the LiDAR data, that cancels the obstacle cost along the medial axis of a narrow passage. A true opening thus becomes a low-cost corridor, while the full repulsive cost is retained everywhere else. The cost reduction factor is embedded at every prediction node of an acados-based nonlinear MPC, which jointly plans the trajectory, yaw, and fold angle of the platform. A user-specified minimum-width limit ensures that the cost reduction factor only opens the corridors that the platform can actually pass. Gazebo/PX4 simulations and hardware experiments on the MorphoCopter demonstrate autonomous folding and unfolding through curved, inclined, and cluttered passages narrower than the unfolded width, where the standard repulsive cost fails to pass safely in most of the trials. The implementation code is available at \href{https://github.com/harshjmodi1996/morphocopter_mpc}{\textcolor{blue}{Github}} and a short video about the paper is available at \href{https://zh.engr.tamu.edu/wp-content/uploads/sites/310/2026/10/Morphocopter_planning_final.mp4}{\textcolor{blue}{\textbf{video link}}}.
\end{abstract}
\begin{IEEEkeywords}
Motion Control, Aerial Systems: Applications, Motion and Path Planning, Morphing Quadrotors
\end{IEEEkeywords}

\section{Introduction}
\label{section:introduction}

Motion planning for quadrotors has been a widely studied subject since the inception of such robots, and the range of missions assigned to them keeps widening \cite{hutto_update}. However, the motion planning through narrow passages remains challenging. Additionally, the morphing quadrotor designs, such as \cite{folding_UAV, harsh_morphocopter}, add to the challenge of such planning with varying drone size and dynamics. A variety of methodologies have been proposed to address these challenges.

Two challenges make this problem hard. First, a morphing platform must plan its trajectory, orientation, and shape together, as the feasible footprint depends on the fold angle. Second, the most useful gaps for such a platform are those narrower than its unfolded body. Classical repulsive obstacle costs with safe cutoff distance seal off exactly these gaps, because the cost grows sharply near the walls and the interior of a tight corridor appears as a barrier to the optimizer. We survey these approaches in Section~\ref{section:related_work}. To overcome these issues, our framework contributes the following:

\begin{enumerate}
    \item A framework that plans trajectory, yaw, and fold angle online for a morphing quadrotor using only a 2D LiDAR for obstacle sensing, enabling autonomous folding and unfolding through passages narrower than the unfolded width.
    \item A novel cost reduction factor $\mu$ that shapes the obstacle cost instead of measuring the clearance. Computed in closed form from the LiDAR data, it cancels the cost along the medial axis of a passage, so a true opening becomes a low-cost corridor.
    \item A user-specified minimum-width limit tying $\mu$ to the morphing geometry, so that only the corridors the folded platform can pass are opened.
    \item Validation in Gazebo/PX4 simulations and on MorphoCopter hardware in curved, inclined, cluttered, and unknown-obstacle scenarios, benchmarked in simulations over repeated trials against a standard APF repulsive cost.
\end{enumerate}

The rest of the article is organized as follows: Section II reviews the related work, Section \ref{section:obstacle_avoidance_cost} introduces the novel obstacle avoidance cost, Section \ref{section:implementation_framework} describes the implementation on morphing quadrotors, Section \ref{section:results} presents the simulation and experiment results, and Section \ref{section:conclusion} concludes the article.

\begin{figure}[t]
    \centering
    \includegraphics[width=0.55\linewidth]{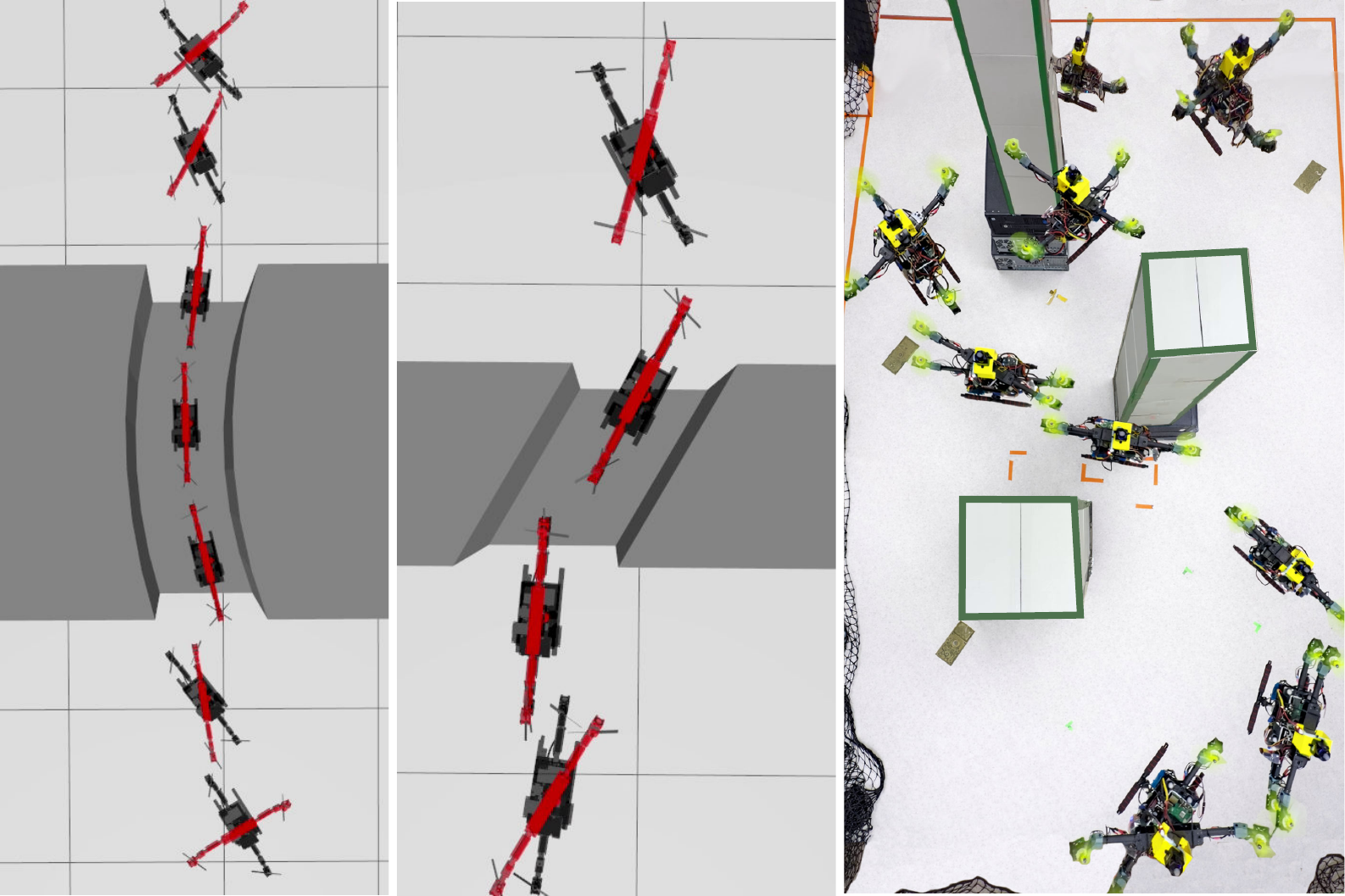}
        
    \caption{MorphoCopter autonomously navigating through ultra-narrow passages using proposed MPC-based motion planning in simulations and experiments}
    \label{fig:overview_figures}
    
\end{figure}
\vspace{-6pt}

\section{Related Work}
\label{section:related_work}
Model-free methods plan and execute trajectories using reinforcement learning or neural networks \cite{morphing_RL, morphocopter_planning_drl_tilted_gap, morphocopter_narrow_gap_3}, but they typically lack explicit dynamic constraints or stability guarantees, and the narrow-gap variants target fixed-frame vehicles with gaps larger than the platform. Optimization and search based planners for fixed-frame drones instead enforce dynamic feasibility through aggressive single-gap flight from onboard vision \cite{morphocopter_narrow_gap_1}, kinodynamic B-spline candidates with collision pruning \cite{morphocopter_planning_fixed_fast_autonomous_trajectory, morphocopter_planning_fixed_fast_autonomous_trajectory2}, point-cloud safe corridors \cite{morphocopter_planning_safety_corridor_point_cloud}, or disturbance-robust narrow-gap traversal \cite{morphocopter_planning_narrow_gap_disturbances_fixed_frame}. Abro et al. \cite{abro_update} apply fractional-order control to trajectory tracking in quadrotor swarms. All assume a fixed footprint, so the gap must exceed the vehicle size.

\begin{table}[htbp!]
    \caption{Comparison with Closest Related Work}
    \label{tab:related_work}
    \vspace{-5pt}
    \setlength{\tabcolsep}{1pt}
    
    \begin{tabular}
    {@{}|m{0.12\textwidth}|m{0.08\textwidth}|m{0.045\textwidth}|m{0.085\textwidth}|m{0.11\textwidth}|@{}}
        \hline
        \rowcolor{gray!30}
        \centering \textbf{Methodology} &
        \centering \textbf{Sensing/ Map} &
        \centering \textbf{Online} &
        \centering \textbf{Passage Assumption} &
        \centering \centering \arraybackslash  \textbf{Key Limitation}\tabularnewline

        \rowcolor{white}
        \hline
         \centering Squeeze drone\cite{folding_UAV} &
         \centering Vision & 
         \centering Partial & 
         \centering 1 known gap &
         \centering \arraybackslash Pre-defined gap\\

         \rowcolor{gray!20}
         \centering Fixed-frame \cite{morphocopter_narrow_gap_1,morphocopter_planning_narrow_gap_disturbances_fixed_frame}  &
         \centering Vision/ LIDAR &
         \centering Yes &
         \centering None &
         \centering \arraybackslash Gap $>$ Vehicle, No morphing
         
         \\

         \rowcolor{white}
         \centering B-spline/ corridor \cite{morphocopter_motion_configuration_control, morphocopter_folding_UAV5} &
         \centering Pre-mapped &
         \centering No &
         \centering Known map &
         \centering \arraybackslash  No replanning \\

         \rowcolor{gray!20}
        \centering Morphing MPC \cite{morphocopter_planning_MPCMDPI, morphocopter_planning_shape_adaptive_astar}&
        \centering Pre-mapped &
        \centering No &
        \centering Known map &
        \centering \arraybackslash  Static env. \\

        \rowcolor{white}
        \centering Geometry-aware NMPC\cite{morphocopter_planning_NMPC_narrowgaps} &
        \centering Known geometry &
        \centering Partial &
        \centering Cubic/ cylindrical &
        \centering \arraybackslash Parametric gap \\

        \rowcolor{gray!20}
        \centering \textbf{Ours} &
        \centering \textbf{2D LiDAR} &
        \centering \textbf{Yes} &
        \centering \textbf{None} &
        \centering \arraybackslash \textbf{2D planning} \\
        \hline
    \end{tabular}
\end{table}

For morphing drones, the survey paper \cite{additional_lou} summarizes various designs, control methodologies and motion planning strategies. B-spline and corridor-based planners \cite{morphocopter_motion_configuration_control, morphocopter_folding_UAV5} and a low-level morphing attitude controller \cite{morphocopter_planning_MPC_lowlevel} address shape change, but they either rely on a pre-built map without replanning, or omit obstacle avoidance. MPC is well suited to such platforms, as it can explicitly consider control and environment constraints over a receding horizon. Chance-constrained MPC plans around moving obstacles for fixed-frame drones \cite{morphocopter_planning_chance_constrained_MPC}. The morphing-specific MPC schemes closest to ours \cite{morphocopter_planning_MPCMDPI, morphocopter_planning_shape_adaptive_astar, morphocopter_planning_NMPC_narrowgaps} either assume static, pre-mapped environments and replan offline \cite{morphocopter_planning_MPCMDPI, morphocopter_planning_shape_adaptive_astar}, or handle only well-defined cubic and cylindrical gaps \cite{morphocopter_planning_NMPC_narrowgaps}. In contrast, our framework replans online at every MPC step directly from raw 2D LiDAR, and the cost shaping makes no assumption on the gap shape. Table \ref{tab:related_work} compares our method with the closest approaches.

Most obstacle-avoidance costs derive from the artificial potential field (APF) \cite{khatib_APF}, whose repulsive terms seal off passages narrower than the activation radius \cite{Koren_APF_narrow_issue, Tilove_APF_narrow_issue}. Control-barrier and signed-distance constraints \cite{Wang_MPC_CBF, Ali_MPC_CBF} restore the safety guarantees, but add non-convex constraints and computational cost. A more recent line of work embeds differentiable collision detection in trajectory optimization: DCOL \cite{dcol} solves a convex program for the minimum scaling that brings two convex shapes into contact, and SILICO \cite{silico} and MOTT \cite{mott} convert this into single-level or analytic conditions for better solver convergence. These methods are rigorous and differentiable, but they model the obstacles as volumetric primitives and solve an inner problem for each shape pair.

More importantly, like the signed-distance and APF costs, these metrics measure proximity, so as a cost they still penalize the interior of a feasible narrow gap. Our cost reduction factor instead shapes the cost in closed form directly from the 2D LiDAR line segments, so it is evaluated at every MPC prediction node without any inner optimization or volumetric reconstruction of the scan. The anti-parallel condition that triggers $\mu$ mirrors the stationarity geometry of the minimum-scaling problem in DCOL \cite{dcol}, and its zero-cost locus coincides with the generalized Voronoi diagram \cite{voronoi}. Hence $\mu$ is a closed-form surrogate for the same medial-axis geometry, suited to per-node evaluation in an online MPC.
\vspace{-6pt}

\section{Obstacle Avoidance Cost}
\label{section:obstacle_avoidance_cost}
The proposed obstacle avoidance cost is computed directly from the 2D LiDAR data, in the horizontal scan plane of the sensor: the line segments and the cost reduction factor are evaluated in 2D, while the altitude is tracked separately by the MPC. For obstacles with 3D non-uniform geometry, such as overhangs or tilted structures, our planner avoids the specific cross-section intersected by the scan plane at the platform's current altitude, but has no explicit reasoning about geometry above or below this plane. So it cannot exploit the clearance that may be available at other altitudes. The extension to volumetric obstacles is left as future work (Section~\ref{section:conclusion}).

\noindent The proposed cost function is of the form:
\begin{small}
    \begin{equation}
        \label{eq:obstacle_avoidance_cost}
        J_o = W_o \cdot (1-(\mu^2-1)^2)\cdot \exp\left(1-\frac{d^{\star^2}}{d^2_0}\right)
    \end{equation}
\end{small}
\noindent Here $W_o \in \mathbb{R}$ is a tunable weight parameter, $d^{\star}$ is the distance to the closest detected obstacle from the prediction location, and $d_0$ is the distance where the cost equals $W_o$. $\mu$ is the novel cost reduction factor responsible for reducing the cost in narrow passages, discussed in the next subsection. The cost term does not have any sharp cutoff, resulting in a smoother gradient throughout the region, unlike standard APF-like cost functions. The factor $(1 - (\mu^2 - 1)^2)$ is flat at $\mu = 0$ and rises monotonically to 1 at $\mu = 1$. The flat bottom keeps the cost near zero across the width of a genuine medial channel rather than only at a single point, so the MPC sees a corridor and not a knife-edge.

\vspace{-10pt}
\subsection{Cost Reduction Factor}
Intuitively, $\mu$ answers a simple geometric question at each predicted position: is this point between two walls that face each other, or is it close to a single wall? $\mu$ approaches 0 when the closest obstacle on one side is balanced by an opposing obstacle on the other side, i.e., on the medial axis of the passage. $\mu$ approaches 1 when the obstacles lie only on one side. Hence, as per (\ref{eq:obstacle_avoidance_cost}), the obstacle cost collapses to zero along the medial axis of a passage and remains unchanged near isolated walls.

\begin{figure}[htbp]
    \centering
    \includegraphics[width=0.8\linewidth]{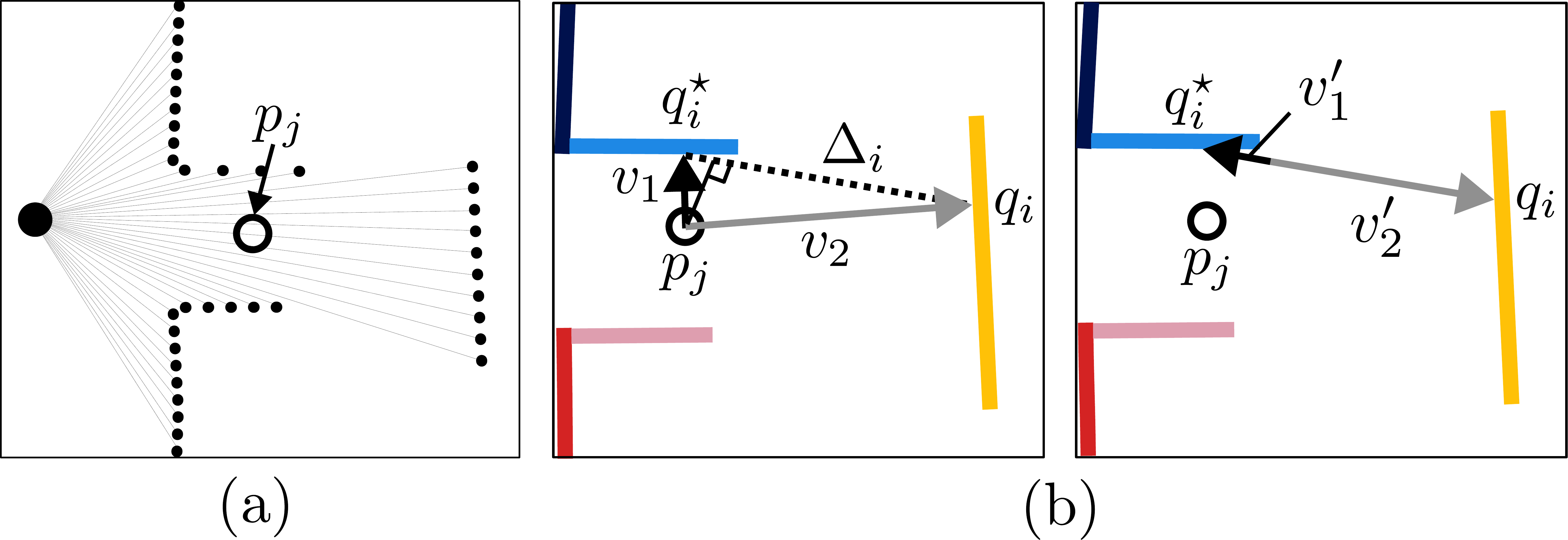}        
    \caption{(a) 2D LiDAR scan (b) Cost reduction factor calculation (see Algorithm \ref{alg:crf_calculation})}
    \label{fig:crf_explanation_new_less_figures}
    
\end{figure}

The factor also resolves a tuning conflict in (\ref{eq:obstacle_avoidance_cost}). Without $\mu$, the only distance control on the cost is $d_0$: a larger $d_0$ gives a safer barrier but seals off narrow passages, while a smaller one opens the passages but reduces the safety margin around every obstacle. $\mu$ decouples these two effects by lowering the cost only inside genuine passages while keeping a high $d_0$ elsewhere. A minimum gap width, set slightly larger than the folded width of the vehicle, ensures that $\mu$ does not open gaps narrower than the platform can safely pass.

\begin{algorithm}[!htbp]
\caption{Line segment extraction from point cloud}
\label{alg:line_segment_extraction}
\begin{algorithmic}[1]
\REQUIRE Raw 2D LiDAR point cloud $\mathcal{P}$ (scan plane), clustering threshold $\epsilon$, minimum cluster size $\mathcal{N}$
\ENSURE Set of merged line segments $\mathcal{L}$\\
\STATE $\mathcal{D} \leftarrow \mathrm{DBSCAN}(\mathcal{P}, \epsilon, \mathcal{N})$
\FOR{each cluster $C_i \in \mathcal{D}$}
    \STATE Sort points in $C_i$ by original index
    \STATE Detect break points $\mathcal{B}$ based on distance trends and discontinuities
    \STATE Split $C_i$ into sub-clusters $\{S_1, S_2, \ldots, S_m\}$ at $\mathcal{B}$
    \FOR{each sub-cluster $S_k$}
        \STATE Fit a line in the scan plane via linear regression
        \STATE Store its endpoints as a line segment $\boldsymbol{L}_k$
    \ENDFOR
    \STATE Merge adjacent line segments if nearly parallel and spatially close
\ENDFOR
\RETURN $\mathcal{L}$
\end{algorithmic}
\end{algorithm}

The calculation of $\mu$ is explained below using Algorithms \ref{alg:line_segment_extraction} and \ref{alg:crf_calculation}, and Fig. \ref{fig:crf_explanation_new_less_figures}. In pre-processing, the 2D LiDAR point cloud is converted into line segments corresponding to the detected obstacles using Algorithm \ref{alg:line_segment_extraction}. The raw scan is passed to DBSCAN without any prior smoothing or outlier filtering. The clustering threshold ($\epsilon$) is set according to the minimum desired gap width. Setting it too small might spuriously detect a non-existent corridor, and setting it too large might seal off the genuine passage. A value slightly above the minimum width of the platform works well in practice, and the gap-width sweep of Section \ref{subsubsection:edge_case_scenarios} verifies this choice. Another DBSCAN parameter, the minimum cluster size $\mathcal{N}$, ensures that the sparse returns from sensor noise are rejected implicitly. Clusters are further divided if the angle between consecutive points exceeds a threshold ($20^\circ$), which allows curved shapes to be approximated by line segments. Finally, for each cluster, line segments are fitted using linear regression. These segments represent the obstacle boundaries facing the sensor.

\begin{figure}[htbp]
    \centering
    \includegraphics[width=0.8\linewidth]{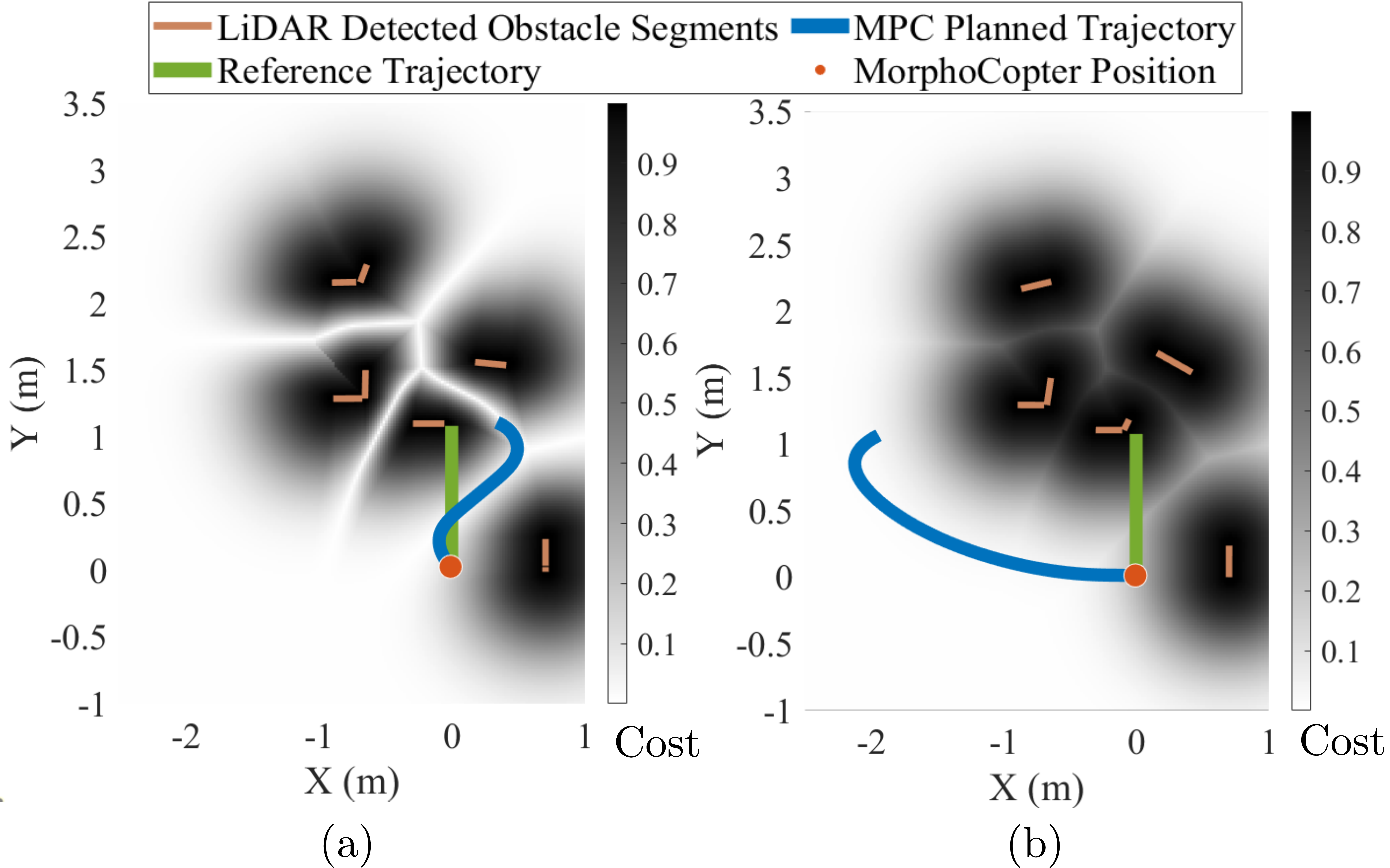}        
    \caption{Comparison of the obstacle avoidance cost function (a) with cost reduction factor and (b) without cost reduction factor. Colormap source \cite{colormaps}}
    \label{fig:crf_nocrf_example}
\end{figure}

The extracted line segments are provided to the acados \cite{acados} MPC loop to calculate the cost reduction factor using Algorithm \ref{alg:crf_calculation}, which can be graphically visualized in Fig. \ref{fig:crf_explanation_new_less_figures}. From each MPC prediction position ($\mathbf{p_j}$), we determine the closest point on each obstacle line segment ($\mathbf{q_i}$) and calculate the vectors ($\boldsymbol{v_i}$) connecting $\mathbf{p_j}$ to the closest points on each line segment. Out of all these vectors, we pair the smallest vector ($\boldsymbol{v_1} = \boldsymbol{v_i^\star}$) with every other vector $\boldsymbol{v_i}$. Now, for each such pair, we project the vectors onto the line ($\mathbf{\Delta_i}$) connecting the respective points on the line segments. Then, we take the normalized vector sum of these projections ($s_i$). The final cost reduction factor for timestep $j$ is the minimum of all such normalized vector sums from all the line segments.

\begin{algorithm}[!htbp]
\caption{Narrow gap cost reduction factor calculation}
\label{alg:crf_calculation}
\begin{algorithmic}[1]
\REQUIRE Line segments $\mathcal{L}$, 2D part of MPC predicted positions $p_j$
\ENSURE Cost reduction factor $\mu_j$ for each MPC prediction timestep $j$

\FOR{each timestep $j$ with predicted position $\mathbf{p_j}$}
    \STATE \COMMENT{Step 1: closest obstacle points}
    \STATE For each segment $i \in \mathcal{L}$, compute the closest point $\mathbf{q_i}$ to $\mathbf{p_j}$ and the distance $\|\mathbf{q_i} - \mathbf{p_j}\|$
    \STATE $i^\star \gets \arg\min_i \|\mathbf{q_i} - \mathbf{p_j}\|$, \ $\mathbf{q^\star} \gets \mathbf{q_{i^\star}}$, \ $\boldsymbol{v_1} \gets \mathbf{q^\star} - \mathbf{p_j}$
    \STATE \COMMENT{Step 2: evaluate narrow-gap interaction}
    \STATE Initialize $\mu_j \gets 1$
    \FOR{each segment $i \in \mathcal{L}$, $i \neq i^\star$}
        \STATE Passage direction $\mathbf{\Delta_i} \gets \mathbf{q^\star} - \mathbf{q_i}$, \ $\boldsymbol{v_2} \gets \mathbf{q_i} - \mathbf{p_j}$
        \STATE Project $\boldsymbol{v_1}, \boldsymbol{v_2}$ onto $\Delta_i$ as $\boldsymbol{v'_1}, \boldsymbol{v'_2}$
        \STATE $s_i \gets \frac{||\boldsymbol{v'_1} + \boldsymbol{v'_2}||}{||\boldsymbol{v'_1}||+||\boldsymbol{v'_2}||}$, \quad $\mu_j \gets \min(\mu_j, s_i)$
    \ENDFOR
\ENDFOR
\RETURN $\{\mu_j\}$
\end{algorithmic}
\end{algorithm}

Consistent with the intuition described earlier, this algorithm yields $\mu_j = 0$ exactly in the middle of a narrow passage, allowing the MPC to plan the trajectory inside it, and $\mu_j = 1$ when the obstacles lie only on one side, leaving the base cost unchanged. Fig. \ref{fig:crf_nocrf_example} shows the effect of $\mu$ on the obstacle cost. Without it, the cost is near its maximum inside the narrow gaps, so planners such as MPC have to route around them. With it, the cost function exposes a low-cost passage for each genuine opening, as elaborated in Section \ref{subsection:simulations}.

\begin{figure}[htbp]
    \centering
    \includegraphics[width=0.55\linewidth]{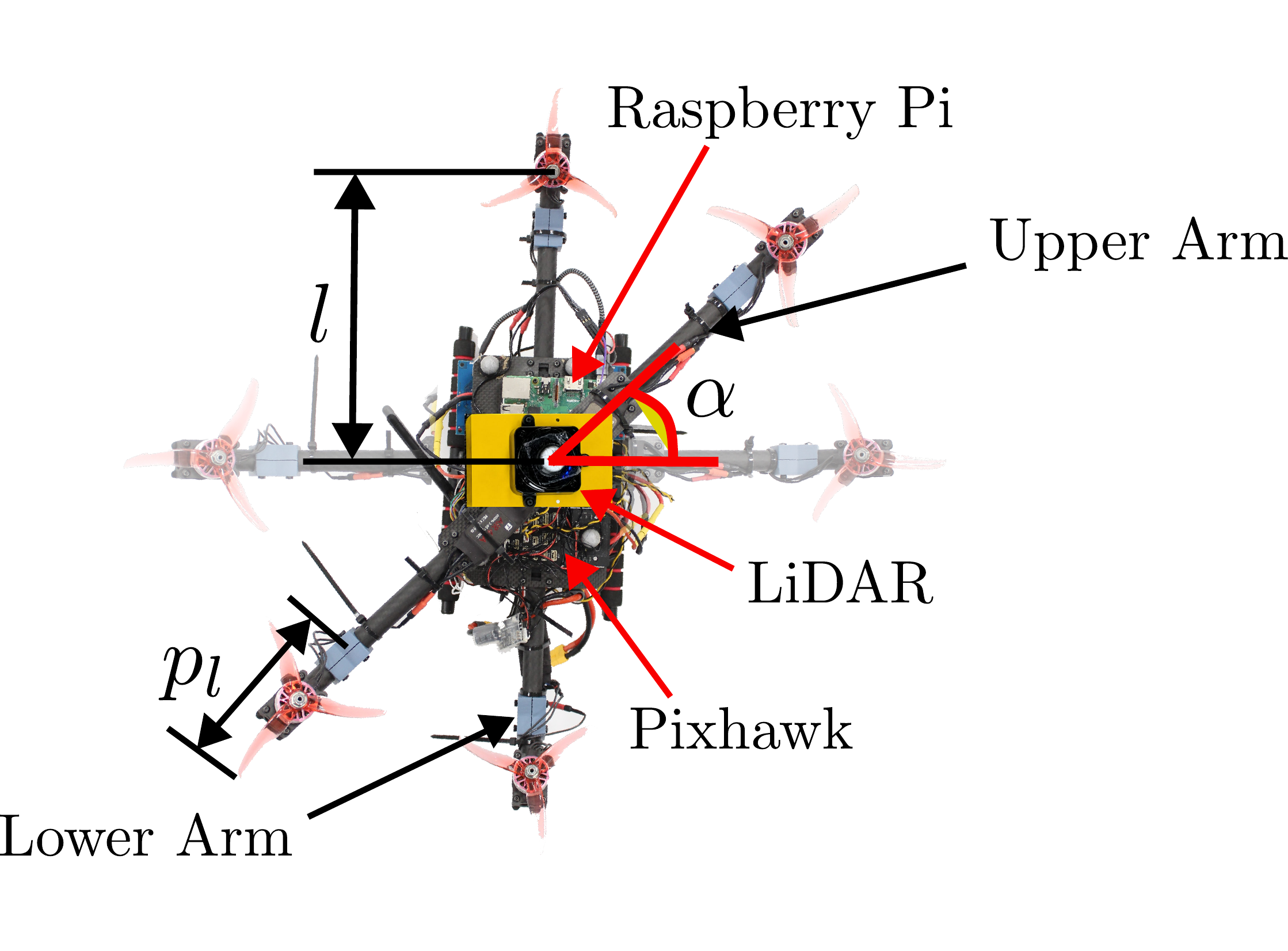}        
    \caption{MorphoCopter hardware details with nomenclature}
    \label{fig:hardware_figure}
\end{figure}

\section{Implementation on Morphing Quadrotors}
\label{section:implementation_framework}

\begin{figure}[htbp]
    \centering
    \includegraphics[width=0.7\linewidth]{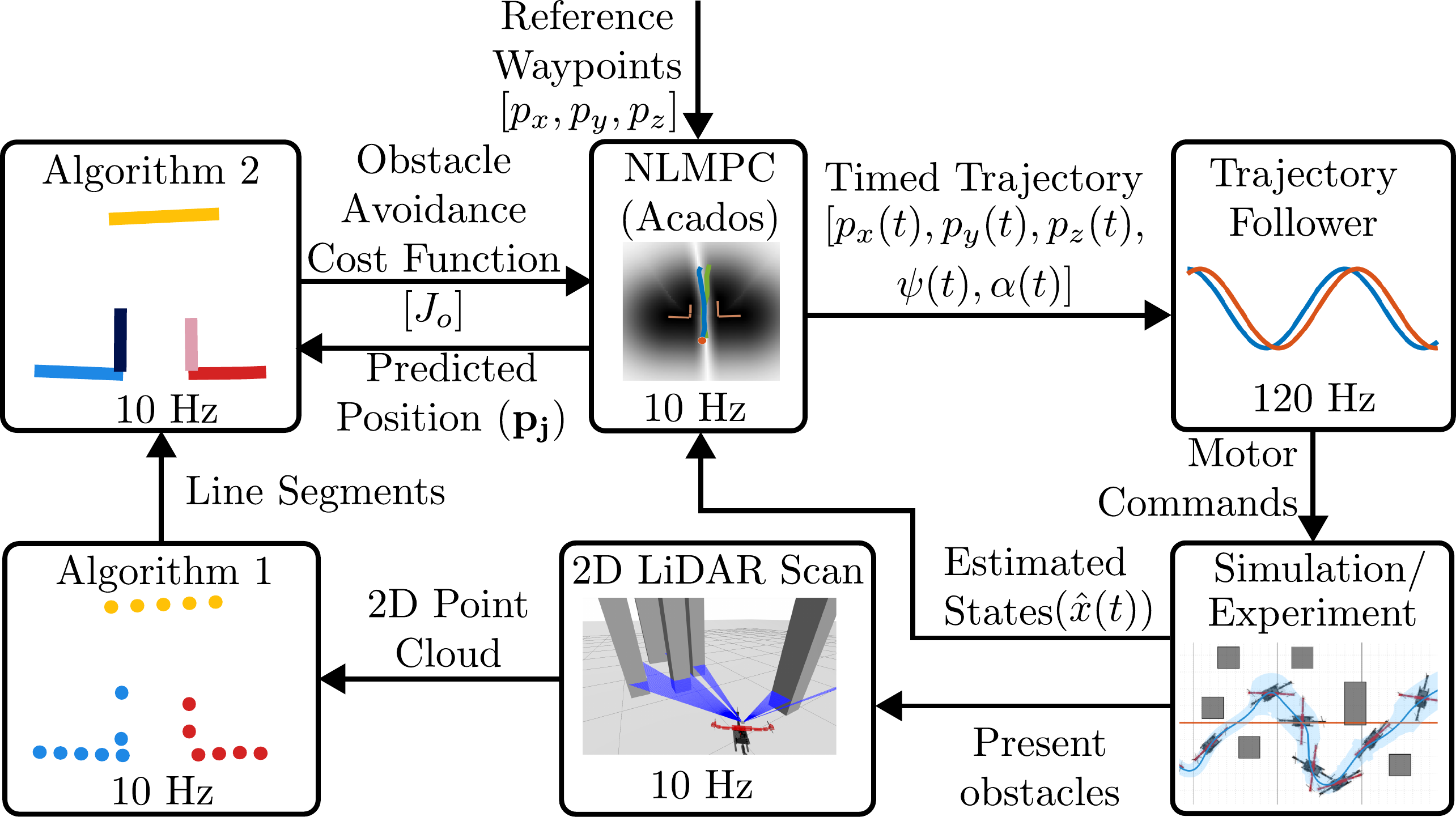}
    \caption{Implementation framework for MorphoCopter MPC. The Nonlinear MPC (NLMPC) jointly optimizes trajectory, yaw, and fold angle using an obstacle cost shaped by the cost reduction factor $\mu$, computed at each predicted position from LiDAR-derived line segments (Algorithms 1–2).}
    \label{fig:framework}
\end{figure}

The formulated cost function is now applied to trajectory planning for morphing quadrotors using MPC. The implementation uses the MorphoCopter\cite{harsh_morphocopter} (Fig.~\ref{fig:hardware_figure}), which can morph in-flight to any configuration ranging from $447\ mm$ to $138\ mm$ wide. Its two arms rotate with respect to each other about an actuated central pivot, and each arm carries two rotors at its tips that provide upward thrust and control the attitude through their speeds. The arms are separated vertically, so the rotors of the two arms do not collide when the vehicle folds. The morphing angle $\alpha$ is zero in the standard X-quadrotor configuration and $\pi/2$ when the platform becomes two bicopters stacked on top of each other. This range enters the MPC directly as the state constraint in (\ref{eq:ocp}). The distance from the center of mass to the propeller center is denoted by $l$. Further details of the structure and the low-level controller are available in \cite{harsh_morphocopter}.

The MPC implementation framework is summarized in Fig. \ref{fig:framework}. The MPC (10 Hz) takes reference waypoints as input, linearly interpolated for each MPC timestep according to the nominal velocity. The reference can be generated by a high-level planner such as RRT or BIT$^*$ \cite{BIT_star} from a stored static map, but this is not a requirement. The low-level trajectory follower \cite{harsh_morphocopter} (120 Hz) tracks the MPC trajectory and keeps the vehicle stable if an MPC update is delayed. The system dynamics, the objective function, and the reference trajectory are discussed next.
\vspace{-10pt}

\subsection{System Dynamics}
The MPC for the MorphoCopter has 14 states as described in Table \ref{tab:states}. Compared to a standard quadrotor MPC, we have two extra states corresponding to the joint angle. Out of these, we consider $p_x$, $p_y$, $p_z$, $\psi$, and $\alpha$ as outputs of the MPC. The inputs are four motor thrusts ($u_1, u_2, u_3, u_4$) and the joint angle torque $u_{\alpha}$.
\vspace{-8pt}
\begin{table}[htpb]
    \caption{MPC States}
    \label{tab:states}
    \centering
    \begin{tabular}{|c|c|c|}
        \hline
        State & Symbol & Description\\
        \hline
        $\boldsymbol{x}(1,2,3)$ & $p_x, p_y, p_z$ & Positions in $x, y, z$ \\
        $\boldsymbol{x}(4,5,6)$ & $v_x, v_y, v_z$ & Velocities in $x, y, z$ \\
        $\boldsymbol{x}(7, 8, 9)$ & $\phi, \theta, \psi$ & Roll, pitch and yaw angles \\
        $\boldsymbol{x}(10, 11, 12)$ & $p, q, r$ & Angular rates about body $x, y, z$ \\
        $\boldsymbol{x}(13, 14)$ & $\alpha, \dot\alpha$ & Joint angle and its rate of change \\
        \hline
    \end{tabular}
    \vspace{4pt}
\end{table}
\noindent Writing $\boldsymbol{p} = \boldsymbol{x}(1{:}3)$, $\boldsymbol{v} = \boldsymbol{x}(4{:}6)$, $\boldsymbol{\Theta} = [\phi,\theta,\psi]^\top$ and $\boldsymbol{\omega} = [p,q,r]^\top$, these states evolve with time as:
\begin{small}
\begin{align}
    \dot{\boldsymbol{p}} &= \boldsymbol{v}, \qquad
    \dot{\boldsymbol{v}} = g\,\boldsymbol{e}_3 - \tfrac{T}{m}\,\mathbf{R}\,\boldsymbol{e}_3 \label{eq:trans_dynamics}\\
    \dot{\boldsymbol{\Theta}} &= \boldsymbol{E}(\boldsymbol{\Theta})^{-1}\boldsymbol{\omega}, \qquad
    \dot{\boldsymbol{\omega}} = \boldsymbol{I}(\alpha)^{-1}\!\left(\boldsymbol{\tau} - \boldsymbol{\omega}\times \boldsymbol{I}(\alpha)\,\boldsymbol{\omega}\right) \label{eq:rot_dynamics}\\
    \dot{\boldsymbol{x}}(13) &= \boldsymbol{x}(14), \qquad
    \dot{\boldsymbol{x}}(14) = \ddot{\alpha} = \frac{u_{\alpha}}{\boldsymbol{I}_{zz}} \label{eq:alpha_dynamics}
\end{align}
\end{small}
\noindent where $g$ is the gravitational constant, $\boldsymbol{e}_3 = [0,0,1]^\top$, $\mathbf{R}$ is the rotation matrix from the body frame to the world frame, $m$ is the MorphoCopter mass, $\boldsymbol{I}(\alpha)$ is the inertia matrix as a function of $\alpha$, $\boldsymbol{\tau} = [\tau_x, \tau_y, \tau_z]^\top$ are the body moments, and $\boldsymbol{E}(\boldsymbol{\Theta})$ is the standard ZYX Euler-rate transformation matrix. $\tau_x$ and $\tau_y$ generated by each motor $i$ depend on $\alpha$ as $\tau_x = \pm[u_i\cos(\delta)(l\sin A) + M_i\sin(\delta) \cos A]$ and $\tau_y = \pm[u_i\cos(\delta)(l\cos A) - M_i\sin(\delta) \sin A]$. Here, $A = (\pi/4-\alpha/2)$, $M_i$ represents the moment generated by motor $i$, and $\pm$ sign depends on the location of the motor. This relationship links system dynamics to $\alpha$. The total thrust along the body $z$ axis is $T = (u_1 + u_2 + u_3 + u_4)\cdot \cos(\delta)$, where $\delta$ is the fixed tilt of the motors as described in \cite{harsh_morphocopter}.
\vspace{-10pt}

\subsection{Objective Function}
The objective function at each stage is $J_j = J_{y,j} + J_{u,j} + J_{o,j}$, where $J_{y,j}$ penalizes deviation from the reference trajectory, $J_{u,j}$ penalizes control inputs deviating from the hover controls, and $J_{o,j}$ is the obstacle avoidance cost. The first two are standard nonlinear least-squares terms:
\begin{small}
\begin{equation}
\label{eq:traj_cost}
    J_{y,j} = \tfrac{1}{2}
\big( \boldsymbol{y}_j - \boldsymbol{y}^{\text{ref}}_j \big)^\top 
\boldsymbol{W}_{y,j}
\big( \boldsymbol{y}_j - \boldsymbol{y}^{\text{ref}}_j \big)
\end{equation}
\end{small}
\noindent Here $\boldsymbol{W}_{y,j} \in \mathbb{R}^{n_y\times n_y}$ is a tunable weight matrix, $\boldsymbol{y}_j$ is the output state of the MPC at stage $j$, and $\boldsymbol{y}_j^{\text{ref}}$ is the reference output state. $J_{u,j}$ is defined analogously with the weight matrix $\boldsymbol{W}_{u,j} \in \mathbb{R}^{n_u\times n_u}$ on $\big( \boldsymbol{u}_j - \boldsymbol{u}^{\text{ref}}_j \big)$, where $\boldsymbol{u}_j^{\text{ref}}$ corresponds to the hover controls in our case. The obstacle avoidance cost $J_{o,j}$ is (\ref{eq:obstacle_avoidance_cost}) evaluated at stage $j$, i.e., with $\mu_j$ and $d_j^{\star}$.
\vspace{-10pt}

\subsection{Reference Trajectory}
The externally provided reference waypoints are linearly interpolated for each MPC stage considering the nominal velocity and fed into the positional reference of $\boldsymbol{y}^{\text{ref}}_j$ in (\ref{eq:traj_cost}). For the yaw and joint angle references, we utilize the MPC trajectory generated in the previous MPC iteration:

\subsubsection{Yaw Reference} 
\begin{small}
    \begin{equation}
        \psi_{j}^{\text{ref}} = \tan^{-1}\left(\frac{p_{y_{j+1}}-p_{y_{j}}}{p_{x_{j+1}}-p_{x_{j}}}\right) \text{ for } 1 < j \leq n-1
    \end{equation}
\end{small}
\noindent and for $j = n$, we consider $\psi_n^{\text{ref}} = \psi_{n-1}^{\text{ref}}$. Here $n$ is the number of timesteps used in the MPC.

\subsubsection{Joint Angle Reference}
For determining the joint angle reference, we first calculate the maximum half-width of the MorphoCopter using $w = \max(d^{\star}_{b_y} - d_{min},\,p_l/2)$. Here, $p_l$ (Fig. \ref{fig:hardware_figure}) is the propeller diameter, $d^{\star}_{b_y}$ is the distance to the closest obstacle projected along the body $y$ axis, and $d_{min}$ is the safety distance desired between the MorphoCopter and the nearest obstacle.

\begin{figure*}[!htbp]
    \centering
    \includegraphics[width=0.9\linewidth]{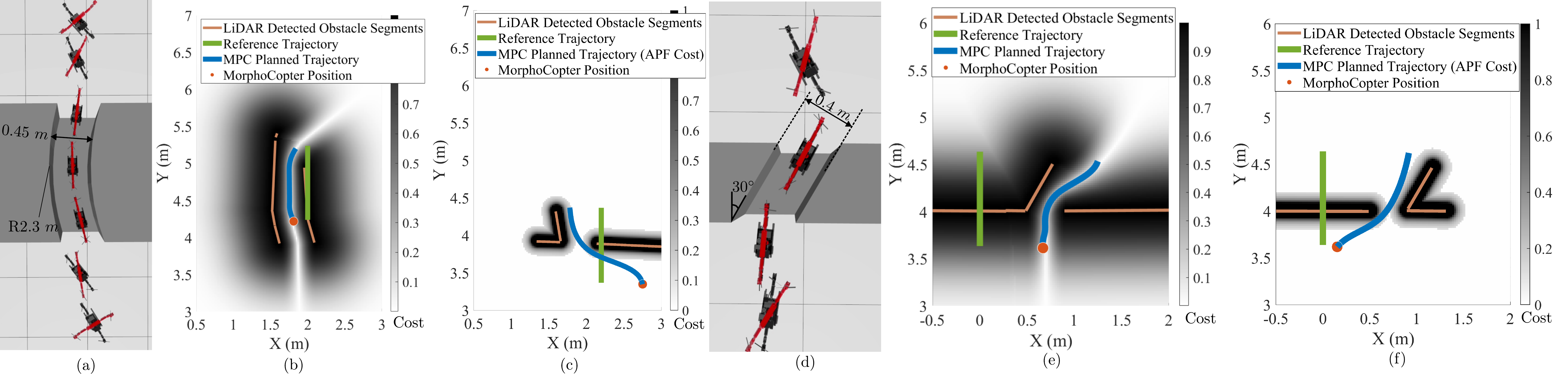}
    \caption{MorphoCopter navigating narrow passages in simulation with the proposed cost function, and the corresponding cost maps for one iteration. (a) Curved passage sequence with proposed cost (b) proposed cost map (c) APF cost map (d) Inclined passage sequence with proposed cost (e) proposed cost map (f) APF cost map.}
    \label{fig: single_wall_sequence}
    
\end{figure*}

Now, if $w > l \cdot \cos(\pi/4)\ +\ p_l/2$, then the reference location does not need any folding and hence $\alpha_j^{\text{ref}} = 0$. Here, $l$ is the distance from the center of mass of the MorphoCopter to the propeller center. If $w \leq l \cdot \cos(\pi/4)\ +\ p_l/2$, we can calculate the reference joint angle using:
\begin{small}
    \begin{equation}
        \alpha_j^{\text{ref}} = \frac{\pi}{2} - 2\cdot \sin^{-1}\left(\frac{w - p_l/2}{l}\right)
    \end{equation}
\end{small}
Note that $\psi^{\text{ref}}_j$ and $\alpha^{\text{ref}}_j$ act only as references. The yaw and the fold angle are genuine MPC states with their own dynamics (\ref{eq:alpha_dynamics}) and constraints. Hence, the MPC jointly optimizes the trajectory, yaw, and fold angle over the horizon, and it is free to deviate from these references whenever the dynamics or the obstacles require it.

\vspace{-10pt}
\subsection{Longitudinal and Lateral Cost Weights}
To let the MPC trajectory deviate laterally from the reference while still following its general direction, we use a much lower cost weight in the lateral direction, converted into standard $x$-$y$ weights at each timestep by:
\begin{small}
    \begin{align}
        \begin{split}
        &\boldsymbol{W}^{p_x,p_y}_{y,j} =  \\
            &\begin{bmatrix}
                w_{v} \cdot c^2(\psi_j) + w_{h}\cdot s^2(\psi_j) & (w_{v}-w_{h})\cdot s(\psi_j)c(\psi_j)\\
                (w_{v}-w_{h})\cdot s(\psi_j)c(\psi_j) & w_{v} \cdot s^2(\psi_j) + w_{h}\cdot c^2(\psi_j)\\
            \end{bmatrix}
        \end{split}
    \end{align}
\end{small}
\noindent where $w_v$ and $w_h$ are the tunable longitudinal and lateral cost parameters, $c$ represents $\cos$ and $s$ represents $\sin$.

\subsection{Complete MPC Formulation}
\label{subsection:constraints}
Collecting the above, the optimal control problem solved at every MPC step is
\begin{small}
\begin{equation}
    \label{eq:ocp}
    \begin{aligned}
        \min_{\boldsymbol{x}_{0:n},\,\boldsymbol{u}_{0:n-1}}\ \ & \sum_{j=0}^{n-1}\left(J_{y,j} + J_{u,j} + J_{o,j}\right) + J_{y,n} + J_{o,n}\\
        \text{s.t.}\ \ & \boldsymbol{x}_0 = \hat{\boldsymbol{x}}(t), \quad \boldsymbol{x}_{j+1} = \boldsymbol{f}(\boldsymbol{x}_j, \boldsymbol{u}_j),\\
        & u_{idle} \leq u_{i,j} \leq u_{max}, \quad i = 1,\ldots,4,\\
        & |u_{\alpha,j}| \leq u_{\alpha,max}, \qquad 0 \leq \alpha_j \leq \pi/2
    \end{aligned}
\end{equation}
\end{small}
\noindent where $\hat{\boldsymbol{x}}(t)$ is the current state estimate and $\boldsymbol{f}$ is the discretized form of (\ref{eq:trans_dynamics})-(\ref{eq:alpha_dynamics}). Here $u_{idle}$ is the thrust generated by the motors while idling, $u_{max}$ is the physical thrust limit of the motors, and $u_{\alpha,max}$ is the torque limit of the joint angle servo. The last constraint is the folding range of the hardware, so the fold angle enters the problem twice: as a state with its own actuator dynamics (\ref{eq:alpha_dynamics}), and as a box constraint on that state at every prediction node. Consequently, the morphing profile returned by the solver is always realizable by the platform, and the obstacle cost $J_{o,j}$ shaped by $\mu$ is evaluated on exactly the same predicted positions that these constraints act upon.
\vspace{-8pt}
\vspace{-2pt}
\begin{table}[!htpb]
    \caption{Tunable Parameters Used}
    \label{tab:parameters}
    \centering
    \small
    \setlength{\tabcolsep}{1pt} 
    \begin{tabular}{|c|c|c|}
        \hline
        Description & Parameters & Values\\
        \hline
            MPC Runtime & $n, t_{del}$ & 25, 0.1 s \\
            Trajectory cost & $w_{v},w_{h},\boldsymbol{W}^{z, \psi, \alpha}_{y}$ & $0.53, 0.026, diag(4.0, 2.0, 20.0)$ \\
            Control cost  & $\boldsymbol{W}_u$ & $diag(0.9, 0.9, 0.9, 0.9, 9.5\cdot10^{-5})$ \\
            Obstacle cost & $\epsilon, \mathcal{N}, d_0, W_o$ & $0.2\ m$, $10$, $0.6\ m, 1.0$ \\
            APF baseline & $\eta$, $d_0$ & $0.001$, $0.1/0.15/0.2/1.6\ m$\\
        \hline
    \end{tabular}
\end{table}

\section{Results}
\label{section:results}
We validate the framework in various simulation and experiment scenarios with narrow passages. In some scenarios, reference trajectories pass directly through obstacles. The tunable parameters used are described in Table \ref{tab:parameters}. We also compare the proposed cost function with the well-established standard APF cost function $[1/2\ \eta (1/d^\star - 1/d_0)^2]$ \cite{khatib_APF} as a baseline. Here $\eta$ is the tunable weight parameter, $d^\star$ and $d_0$ are the same as described in (\ref{eq:obstacle_avoidance_cost}). Table~\ref{tab:quant} summarizes computation times, success rate, and trajectory deviation for the simulation scenarios. In summary, in cases where the MorphoCopter requires a large deviation from the reference trajectory to avoid the obstacles, APF-based MPC has a very low success rate, while the proposed cost remains collision-free in at least 7 of 10 trials in every scenario. Aggregated over the 60 trials of the six scenarios in which $\mu$ is active, the proposed cost achieves a significantly higher success rate and a lower collision rate than the APF baseline (Success / Deviation / Collision count of 48 / 5 / 7 for proposed vs 13 / 0 / 47 for APF). The mean solve time stays within the $100\ ms$ MPC period across all scenarios, and LiDAR data processing is close to $5\ ms$. All collisions of the proposed method in Table~\ref{tab:quant} share one failure mode: the obstacle cost is a soft penalty on the predicted center of mass, so no hard constraint acts on the swept footprint of the folded arms. In these runs, the center path stayed inside the passage while a propeller tip clipped a corner, mostly at a shallow entry angle or before the yaw and fold angle had converged. A hard footprint constraint \cite{Wang_MPC_CBF, dcol} could address this, but it adds a non-convex constraint per node and may render the problem infeasible inside a tight gap under noisy perception. For better visualization of the implementation, please refer to the accompanying video.
\vspace{-10pt}

\subsection{Simulations}
\label{subsection:simulations}
The simulations are performed in the high-fidelity simulator Gazebo using PX4 firmware. A virtual LiDAR sensor is added on top of the MorphoCopter to detect obstacles in the environment. In most cases, straight path references are considered to make the scenarios more challenging.

\begin{figure}[!htbp]
    \centering
    \includegraphics[width=0.7\linewidth]{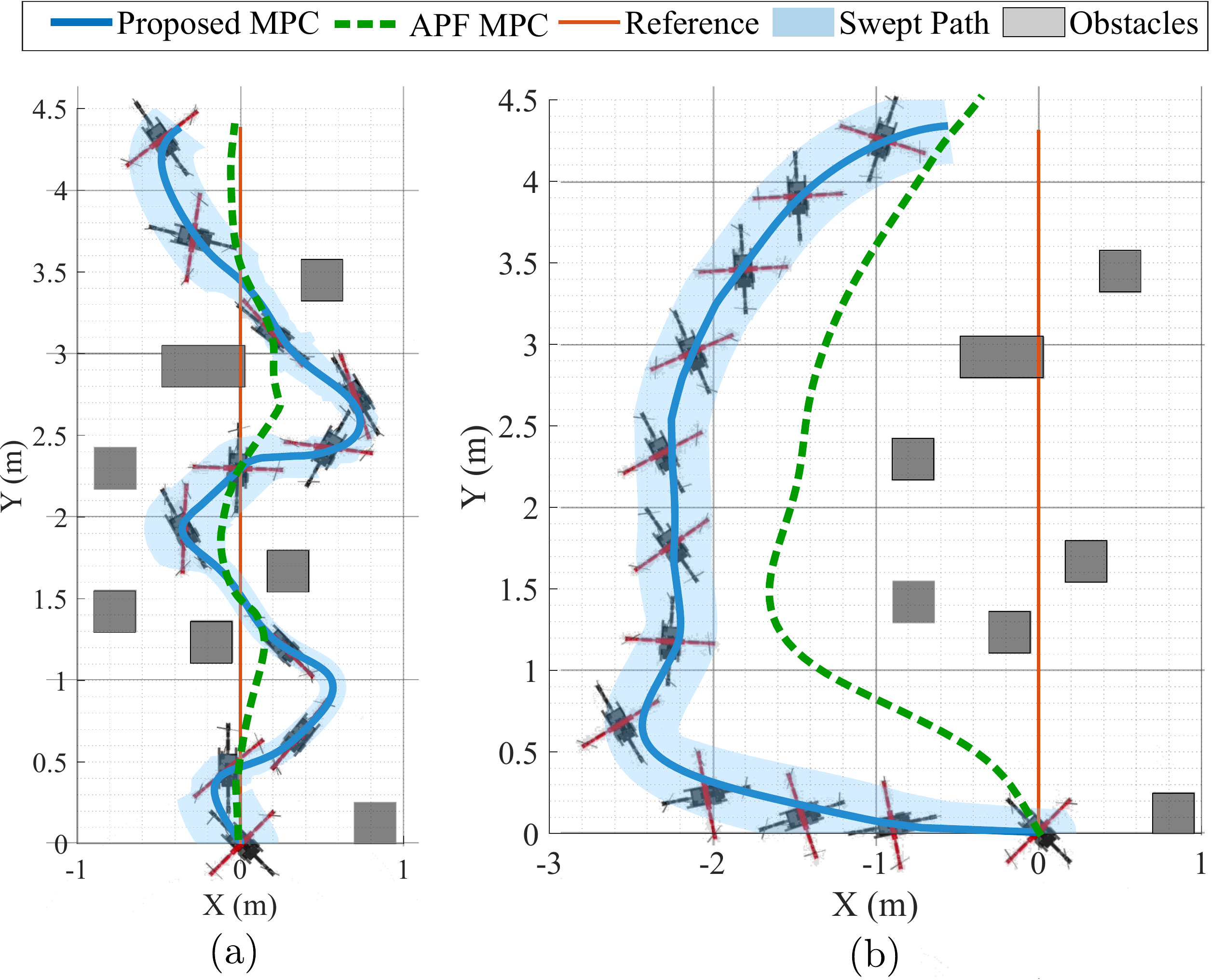}
    \caption{MorphoCopter navigating through a cluttered environment in simulations using (a) the proposed cost function and (b) the same cost without the reduction factor. Dashed lines show the APF baseline trajectory.}
    \label{fig: sim_cluttered_sequence_and_plot}
    \vspace{-5pt}
\end{figure}

\subsubsection{Curved and inclined narrow passages}
In the first simulation scenario, curved and inclined gaps narrower than the unfolded MorphoCopter are generated as shown in Fig. \ref{fig: single_wall_sequence} (a) and (d). The curved passage is $0.45\ m$ wide with a $2.3\ m$ radius, and its curvature forces the swept footprint of the unfolded arms past the walls, so the platform must morph to follow the curve. Fig. \ref{fig: single_wall_sequence} (b) shows the obstacle avoidance cost for one iteration, where the cost reduction factor generates a narrow path along the middle of the gap. Even with limited 2D LiDAR sensing, the cost function is adequate for this challenging environment. Fig. \ref{fig: single_wall_sequence} (c) shows the APF cost for the same scene. The activation distance had to be lowered to $0.1\ m$ for the baseline, since a larger value seals off the gap entirely. At such a small activation distance, the cost rises very steeply near the walls, which drives the MPC closer to obstacles and ends in a collision in all 10 trials. All cost maps shown are for illustration only. The MPC evaluates the cost at the predicted positions, not over the whole map.

Similarly, Fig. \ref{fig: single_wall_sequence} (d) and (e) show the MorphoCopter passing through an inclined passage of $0.4\ m$ width and the corresponding cost function. The straight reference was deliberately offset from the passage to demonstrate the ability of the framework to find the safe path through the corridor. The corresponding APF cost map is shown in Fig. \ref{fig: single_wall_sequence} (f). As in the curved case, the steep rise near the obstacles leaves only 1 successful run out of 10, against 7 successful traversals with the proposed cost (Table \ref{tab:quant}).

\subsubsection{Cluttered obstacles}
\begin{table*}[!htbp]
    \caption{Repeated-trial outcomes and computation times for the simulation scenarios. Succ.: passed through the narrow gap; Dev.: reached the goal around the obstacles; Coll.: collided. Lateral RMSE is reported separately for the successful and the deviating runs, and omitted where no such run occurred.}
    \label{tab:quant}
    \centering
    \footnotesize
    \setlength{\tabcolsep}{4pt}
    \begin{tabular}{|l|c|c|c|c|c|c|c|c|c|c|}
        \hline
        \multirow{2}{*}{Scenario} & Gap & \multirow{2}{*}{Ref.} & LiDAR & \multicolumn{2}{c|}{MPC solver (ms)} & APF $d_0$ &\multicolumn{2}{c|}{Succ. / Dev. / Coll.} & \multicolumn{2}{c|}{Lateral RMSE, Succ. / Dev. (m)}\\
        \cline{5-6}
        \cline{8-11}
        & (mm) & & (ms) & Proposed & APF & (m) & Proposed & APF & Proposed & APF\\
        \hline
        Curved passage & 450 & Straight & 4.8 & 50 & 56 & 0.10 & 7 / 2 / 1 & 0 / 0 / 10 & 0.251 / 2.857 & -- / -- \\
        Inclined passage & 400 & Straight & 4.7 & 61 & 62 & 0.20 &  7 / 1 / 2 & 1 / 0 / 9 & 0.578 / 1.594 & 0.248 / -- \\
        Cluttered, with $\mu$ & Var. & Straight & 5.3 & 81 & 58 & 0.20 &  8 / 2 / 0 & 3 / 0 / 7 & 0.444 / 1.881 & 0.175 / -- \\
        Cluttered, no $\mu$ & Var. & Straight & 5.2 & 61 & 42 & 1.60 &  0 / 10 / 0 & 0 / 9 / 1 & -- / 1.848 & -- / 1.035 \\
        Realistic room & 330 & BIT$^*$ & 5.8 & 85 & 72 & 0.15 &  7 / 0 / 3 & 8 / 0 / 2 & 0.322 / -- & 0.320 / -- \\
        Realistic room + unknown obstacle & 330 & BIT$^*$ & 5.9 & 81 & 84 & 0.15 &  9 / 0 / 1 & 1 / 0 / 9 & 0.649 / -- & 0.139 / -- \\
        Perpendicular obstacles & 210 & Straight & 4.8 & 64 & 63 & 0.10 &  10 / 0 / 0 & 0 / 0 / 10 & 0.279 / -- & -- / -- \\
        \hline
    \end{tabular}
\end{table*}

Here, multiple obstacles are placed very close to each other, creating various passages narrower than the unfolded width of the MorphoCopter, with a straight-line reference trajectory. Fig. \ref{fig: sim_cluttered_sequence_and_plot} (a) shows the MorphoCopter navigating this environment along with the swept path, which covers its envelope considering the folding angle. With the proposed framework, it navigates through this environment comfortably while folding as needed. For benchmark purposes, the same simulation without the cost reduction factor is shown in Fig. \ref{fig: sim_cluttered_sequence_and_plot} (b): the MPC cannot find a path between the obstacles and has to plan around them, giving a lateral RMSE of $1.848\ m$ against only $0.444\ m$ with $\mu$ (averaged over the 10 trials). The corresponding cost functions were shown earlier in Fig. \ref{fig:crf_nocrf_example}. 

We also simulate with the APF baseline, with its trajectory shown as a dashed line in both plots. The activation distance is set to $0.2\ m$ when comparing against the proposed cost with $\mu$, and to $1.6\ m$ when comparing against the case without $\mu$, so that each pair of planners is compared at a matched level. In the first case (Fig. \ref{fig: sim_cluttered_sequence_and_plot} (a)), the steep cost gradient close to the obstacles forces the baseline very close to obstacles, and its success rate drops to 3 out of 10 against 8 out of 10 for the proposed cost. In the second case, both planners detour around the whole cluster. The baseline detour is the tighter of the two ($1.035\ m$ against $1.848\ m$) because its repulsion decays faster away from the obstacles, but that smaller margin also produces the one collision recorded in Table \ref{tab:quant}.

\begin{figure}[!htbp]
    \centering
    \includegraphics[width=0.75\linewidth]{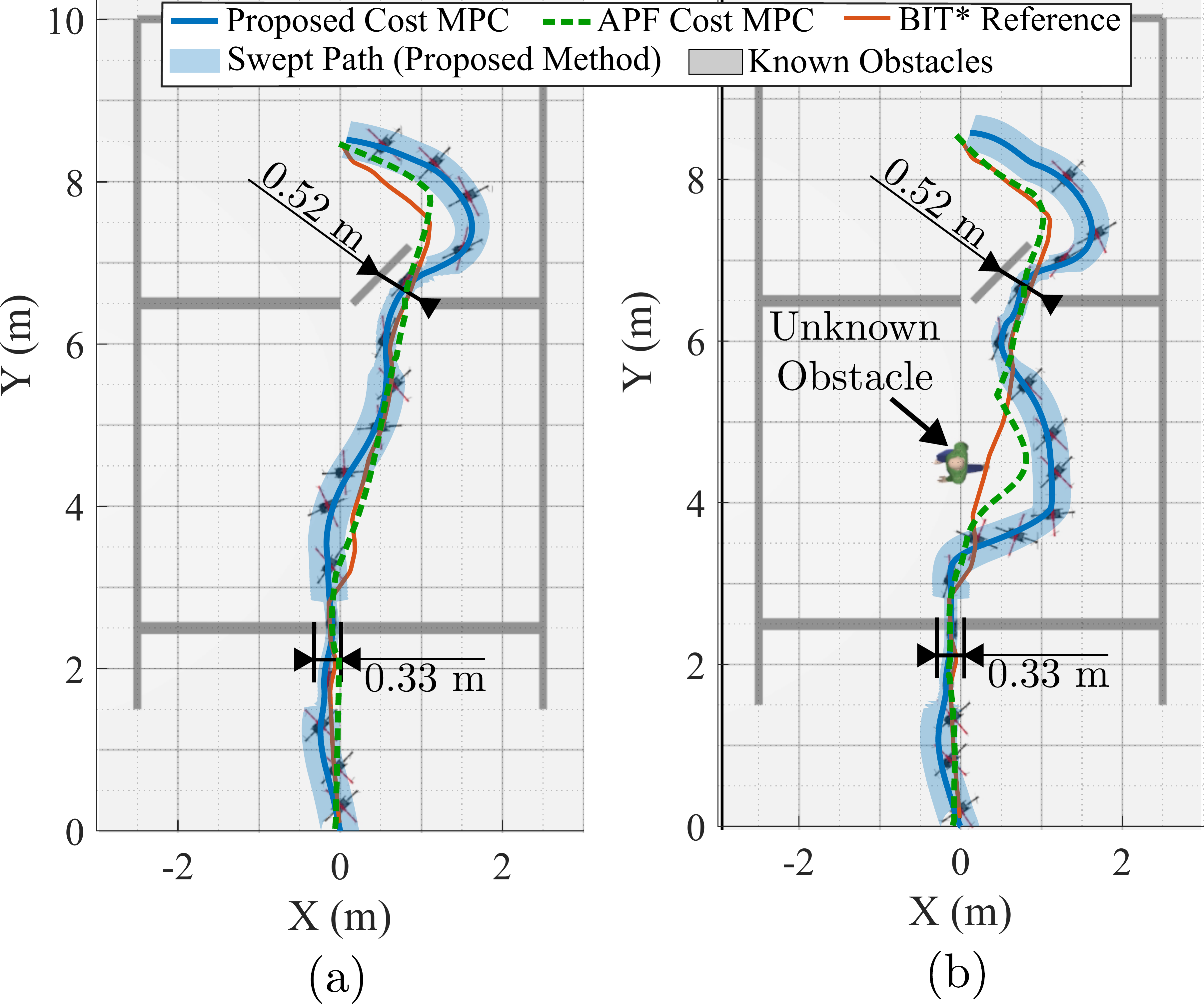}
    
    \caption{MorphoCopter navigating through a realistic room environment in simulations (a) without an unknown obstacle (b) with an unknown obstacle.}
    \label{fig: sim_realistic_room_human_no_human_swept_paths}
    
\end{figure}

\subsubsection{Realistic room}
In this simulation scenario, a realistic room with a window and an ajar door was constructed as shown in Fig. \ref{fig: sim_realistic_room_human_no_human_swept_paths}. The window creates a narrow gap with a width of $ 0.33 \ m$, and the ajar door creates an inclined narrow passage of $0.52\ m$ width. In this simulation, we utilize BIT$^*$ to generate the reference trajectory. Fig. \ref{fig: sim_realistic_room_human_no_human_swept_paths} (a) shows the obstacles, the BIT$^*$ reference, and the MPC trajectories for both the proposed and baseline cost functions. The figure also shows the swept path and simulation snapshots for the proposed cost function case. The BIT$^*$ trajectory passes very close to obstacles at multiple locations, but the MPC with the proposed cost plans a smooth and dynamically feasible trajectory to avoid them. The baseline trajectory passes closer to the obstacles than the trajectory with the proposed cost.  

Fig. \ref{fig: sim_realistic_room_human_no_human_swept_paths} (b) has one unknown obstacle, through which the BIT$^*$ trajectory is passing, as the obstacle was not present while generating the reference trajectory. The MPC with the proposed cost plans a smooth trajectory around the new obstacle. The replanned trajectory does not trigger morphing as the available space is still large enough to accommodate the unfolded configuration. The low activation distance again keeps the baseline trajectory close to the obstacles, and only 1 of its 10 runs is collision-free, against 9 out of 10 with the proposed cost. We also tried the same scenario with the moving unknown obstacle, which is one of the failure modes of the proposed methodology. As the cost is recomputed from the current scan at every MPC step, such obstacles are handled reactively rather than predicted ahead of time. An approach fast enough to outpace the 10 Hz replanning rate can lead to a collision. We plan to address this in future work.

\begin{figure}[!htbp]
    \centering
    \includegraphics[width=0.8\linewidth]{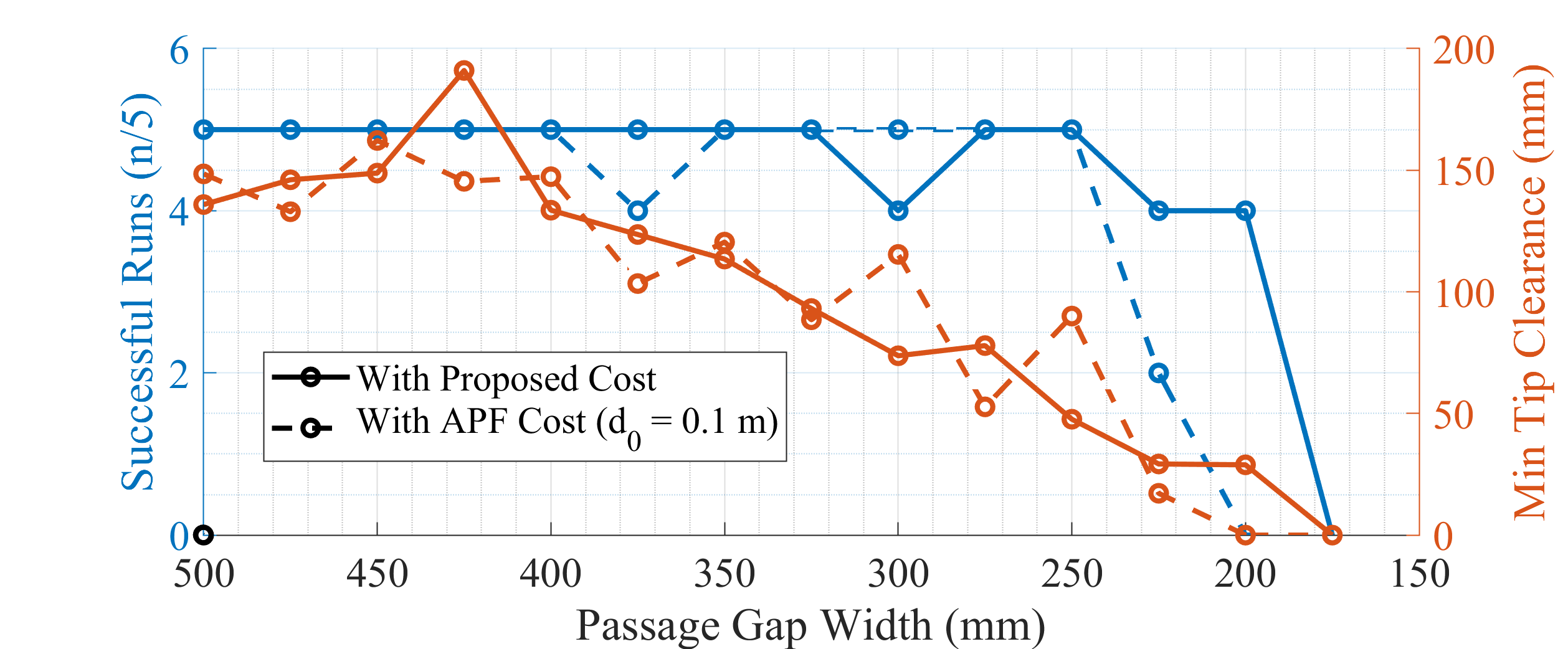}
    \caption{Traversal success rate and minimum propeller-tip clearance for various passage gap widths}
    \label{fig: gap_sweep_plots} 
\end{figure}

\subsubsection{Gap Width Sweep}
\label{subsubsection:edge_case_scenarios}
We now quantify the performance of the framework near its feasibility limit. The passage gap width is swept from $500\ mm$ down to $175\ mm$ in simulation with the minimum-width limit kept at $150\ mm$, and 5 runs are performed at each width. Fig.~\ref{fig: gap_sweep_plots} shows the traversal success rate along with the minimum propeller-tip clearance observed. The success rate remains high up to $200\ mm$, then falls to 0 at $175\ mm$, as this gap is barely larger than the minimum width of the MorphoCopter ($138\ mm$) and leaves a margin of only $18\ mm$ on each side. This might be due to perception noise and tracking errors. We therefore keep the minimum gap-width parameter $(\epsilon)$ at $200\ mm$ for all other simulations and experiments. For APF baseline simulations, the success rate falls sharply at a 225$\ mm$ gap. The tip clearance remains comparable to the proposed case as the reference trajectory and narrow gap align with each other in this scenario.

\subsubsection{Edge Case Scenarios}
\begin{figure}[!htbp]
    \centering
    \includegraphics[width=0.9\linewidth]{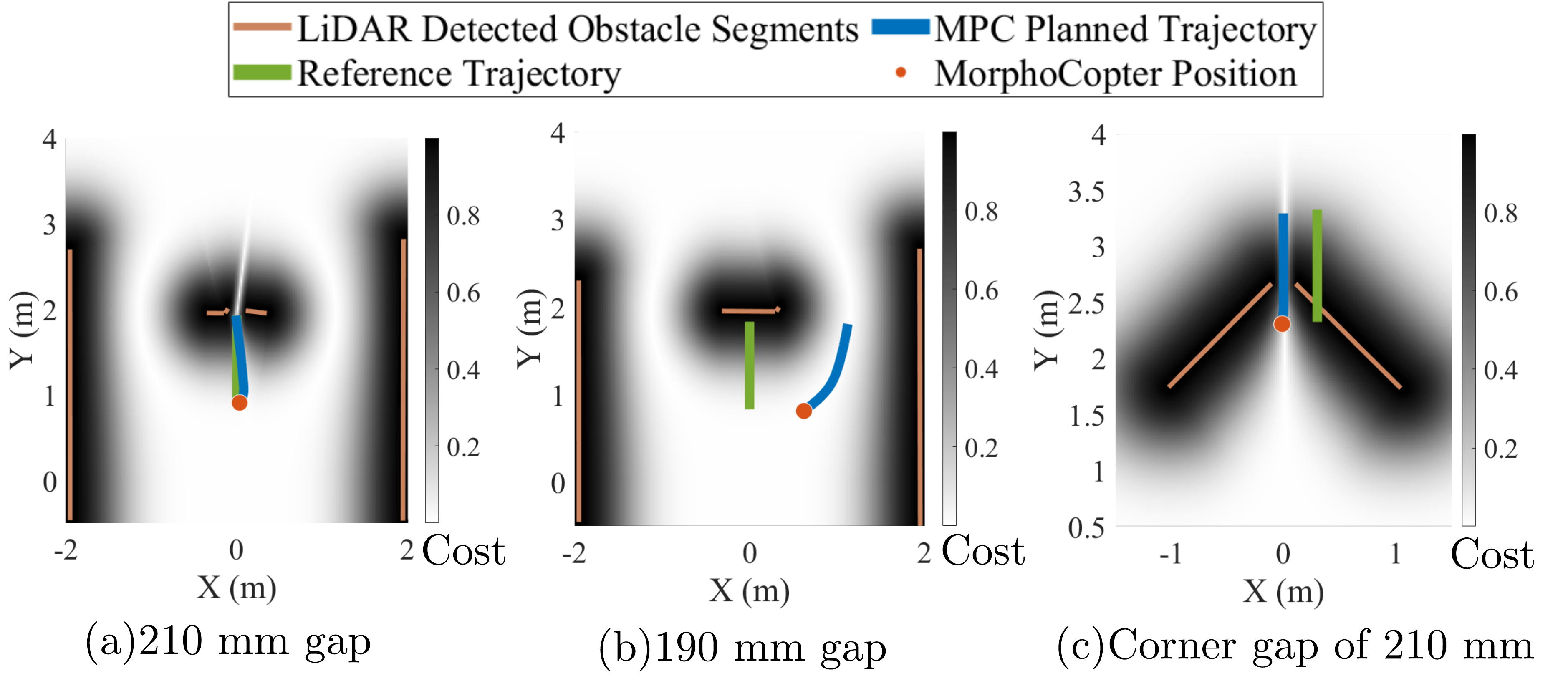}
    \caption{Edge cases. (a) Minimum-width limit of $200\ mm$ with a $210\ mm$ gap: $\mu$ opens a low-cost corridor. (b) Same limit with a $190\ mm$ gap: the faces merge into one cluster and the gap stays sealed. (c) Non-intersecting perpendicular obstacles}
    \label{fig: comparison} 
\end{figure}

We further examine two limiting behaviors of $\mu$ using the cost functions shown in Fig.~\ref{fig: comparison}. The first is around the minimum-width limit. In Fig.~\ref{fig: comparison} (a), a $200\ mm$ limit with a $210\ mm$ gap keeps the two walls in separate clusters, and $\mu$ opens a low-cost channel along the middle. Narrowing the gap to $190\ mm$ in Fig.~\ref{fig: comparison} (b) merges the two faces into a single cluster, so $\mu$ stays at 1, the gap remains fully penalized, and the MPC treats it as a wall.

The second behavior is with non-parallel obstacles. $\mu$ does not require the two faces to be parallel, because it detects an opposing-wall opening rather than a parallel corridor. Fig.~\ref{fig: comparison} (c) shows two non-intersecting perpendicular obstacles: the opening between them is wider than the minimum-width limit, so $\mu$ opens a low-cost passage exactly as it does for parallel walls. The cost shaping thus depends only on the existence and the width of an opening, and not on the orientation of the obstacles forming it.

\begin{figure}[!htbp]
    \centering
    \includegraphics[width=0.75\linewidth]{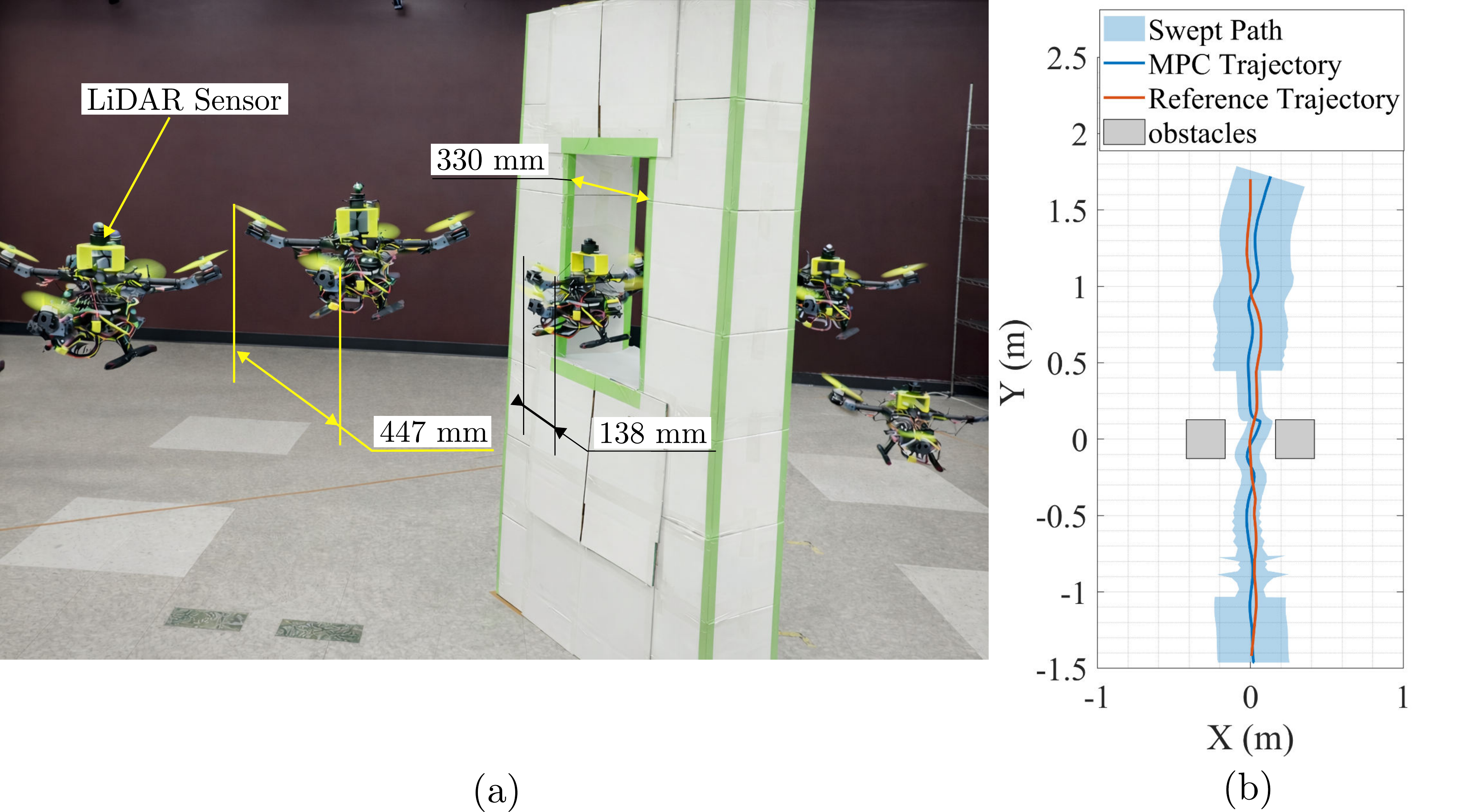}
    
    \caption{MorphoCopter passing through an extremely narrow gap using online MPC trajectory planning with LiDAR perception (a) experiment video sequence (b) trajectories and swept path}
    \label{fig: old_obstacle_sequence_and_swept_path}
    
\end{figure}

\vspace{-10pt}

\begin{figure}[!htbp]
    \centering
    \includegraphics[width=0.75\linewidth]{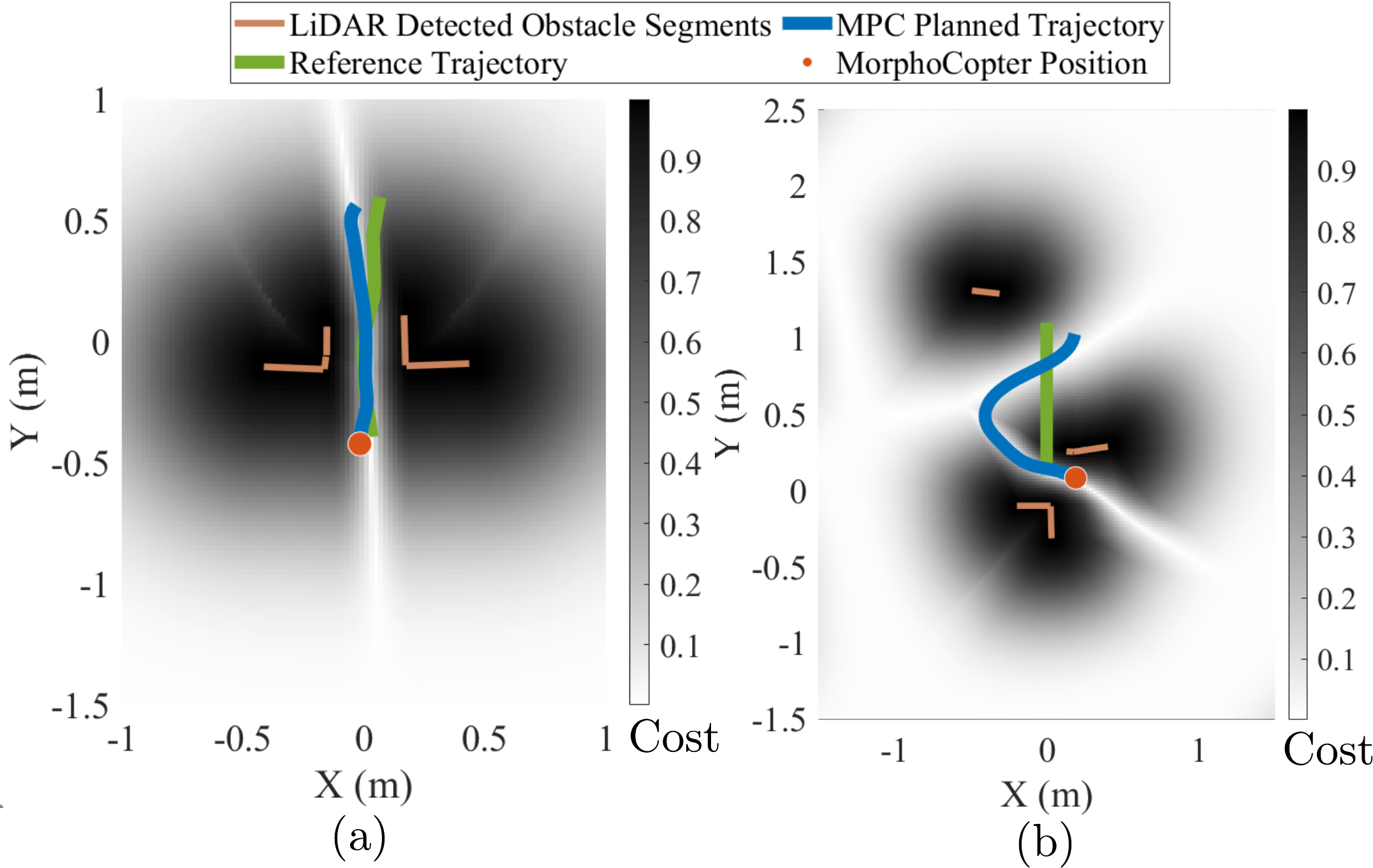}
    \vspace{-5pt}
    \caption{Obstacle avoidance cost function for one of the iterations in experiments (a) single narrow gap environment (b) cluttered obstacles environment}
    \label{fig: cost_both_exp}
    \vspace{-5pt}
\end{figure}

\begin{figure}[!htbp]
    \centering
\includegraphics[width=0.55\linewidth]{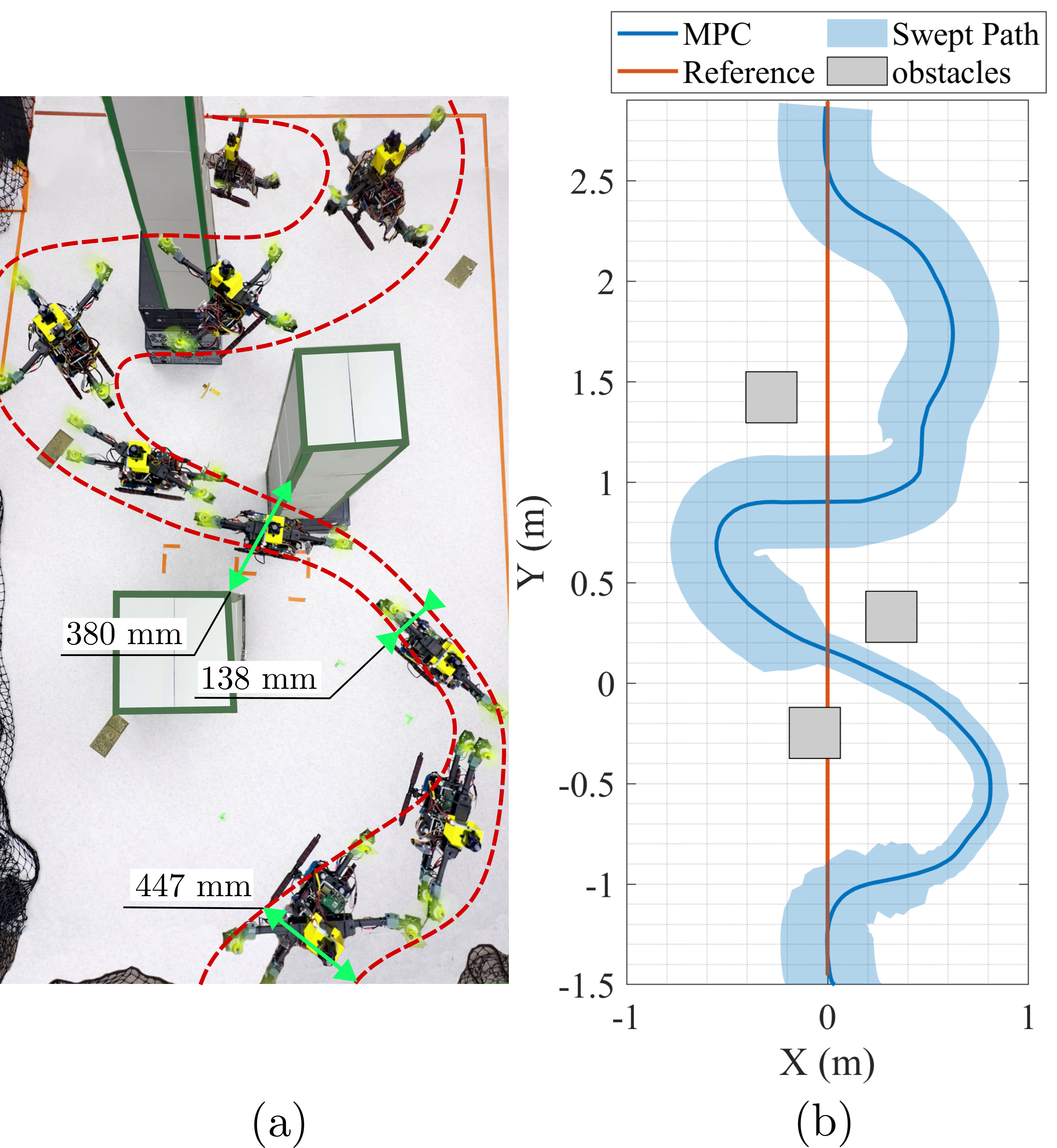} \vspace{-5pt}
    \caption{{MorphoCopter passing through a cluttered obstacles environment using online MPC trajectory planning with LiDAR perception (a) experiment video sequence (b) trajectories and swept path}}
    \label{fig: sequence_with_overlay}
    
\end{figure}

\subsection{Experiments}
We validate the framework on the MorphoCopter hardware \cite{harsh_morphocopter}. As shown in Fig. \ref{fig:hardware_figure}, the setup has an onboard Pixhawk flight controller, a Raspberry Pi, markers for Vicon localization, and, in addition to the design of \cite{harsh_morphocopter}, a 2D LiDAR for onboard obstacle sensing. Experiments are conducted in two environments: a single narrow passage and a cluttered space with tight gaps among the obstacles. The onboard computer runs only the trajectory follower, while the MPC runs offboard and streams its trajectory over Wi-Fi. The measured round-trip communication latency stayed below $5\ ms$, which is negligible against the $100\ ms$ MPC period. Also, the $120\ Hz$ onboard trajectory follower keeps the vehicle stable on the few occasions when the MPC calculation was delayed. The APF baseline was not flown on hardware to avoid damaging the platform. The hardware trials here demonstrate feasibility under real sensor noise, while the repeated-trial statistics are reported in simulation only as described previously.

\subsubsection{Single Narrow Gap}

A narrow passage of $0.33\ m$ width is constructed as shown in Fig. \ref{fig: old_obstacle_sequence_and_swept_path} (a), where the MorphoCopter passes through safely using the proposed framework. Fig. \ref{fig: old_obstacle_sequence_and_swept_path} (b) shows the reference trajectory, the MPC trajectory, and the swept path, which illustrates that the platform folds at a safe distance before entering the gap and unfolds just after clearing it. Fig. \ref{fig: cost_both_exp} (a) shows the cost function for one iteration. The line segments and the cost are generated properly even with noisy hardware measurements, indicating the robustness of the framework.

\subsubsection{Cluttered Obstacles}
Here, three obstacles are arranged with narrow gaps among them, as shown in Fig. \ref{fig: sequence_with_overlay} (a). The narrowest passage is $0.38\ m$ wide and lies at around $45^\circ$ to the reference trajectory. A straight-line reference connects the start and the goal, so the MPC is fully responsible for planning the trajectory and the folding while adjusting the yaw and the joint angle simultaneously. Fig. \ref{fig: cost_both_exp} (b) shows the cost function for one iteration, and Fig. \ref{fig: sequence_with_overlay} (b) compares the reference with the MPC trajectory and the swept path. The MPC trajectory deviates by around $0.75\ m$ in the X direction at one point, demonstrating the obstacle cost safely guiding the MorphoCopter among the obstacles.
\vspace{-7pt}

\section{CONCLUSIONS}
\label{section:conclusion}
This article presented an online trajectory and morphology planning framework for morphing quadrotors to traverse passages narrower than their unfolded width using only a 2D LiDAR. The core enabler is the cost reduction factor $\mu$, which cancels the obstacle cost along the medial axis of a genuine passage while retaining the full repulsive cost elsewhere, and a minimum-width limit ties this shaping to the folded geometry of the vehicle so that only feasible corridors are opened. The cost is embedded in an acados-based nonlinear MPC that jointly plans the trajectory, yaw, and fold angle. The framework was validated in Gazebo/PX4 simulations and MorphoCopter hardware experiments, and benchmarked over repeated trials against a standard APF repulsive cost, which fails to pass safely in most of the trials. The current methodology plans only in the 2D scan plane and handles moving obstacles reactively, both of which we plan to address in future work.
\vspace{-10pt}


\end{document}